\documentclass[conference]{IEEEtran}

\usepackage{cite}
\usepackage{amsmath,amssymb,amsfonts}
\usepackage{algorithmic}
\usepackage{graphicx}
\usepackage{textcomp}
\usepackage{xcolor}
\usepackage{comment}
\usepackage{scalerel,stackengine}
\def\BibTeX{{\rm B\kern-.05em{\sc i\kern-.025em b}\kern-.08em
    T\kern-.1667em\lower.7ex\hbox{E}\kern-.125emX}}
\usepackage{eso-pic}
\usepackage{multirow}

\ifCLASSINFOpdf

\else
 
\fi

\usepackage{eso-pic}
\newcommand\AtPageUpperMyright[1]{\AtPageUpperLeft{%
 \put(\LenToUnit{0.5\paperwidth},\LenToUnit{-1cm}){%
     \parbox{0.5\textwidth}{\raggedleft\fontsize{9}{11}\selectfont #1}}%
 }}%
\newcommand{\conf}[1]{%
\AddToShipoutPictureBG*{%
\AtPageUpperMyright{#1}
}
}

\begin{document}

\conf{2019 22nd International Conference on Computer and Information Technology (ICCIT), 18-20 December, 2019}

\title{A Continuous Space Neural Language Model for Bengali Language \\}

\author{
\IEEEauthorblockN{Hemayet Ahmed Chowdhury}
\IEEEauthorblockA{Department of Computer \\Science and Engineering\\
Shahjalal University of\\
Science and Technology\\Sylhet, Bangladesh\\
Email: hemayetchoudhury@gmail.com}
\and
\IEEEauthorblockN{Md. Azizul Haque Imon}
\IEEEauthorblockA{Department of Computer \\Science and Engineering\\
Shahjalal University of\\
Science and Technology\\Sylhet, Bangladesh\\
Email: azizulhaqueimon@gmail.com}
\and
\IEEEauthorblockN{Anisur Rahman}
\IEEEauthorblockA{Department of Computer \\Science and Engineering\\
Shahjalal University of\\
Science and Technology\\Sylhet, Bangladesh\\
Email: emailforanis@gmail.com}
\and
\hspace{3cm}
\IEEEauthorblockN{Aisha Khatun}
\hspace{3cm}
\IEEEauthorblockA{\hspace{3cm}Department of Computer \\\hspace{3cm}Science and Engineering\\\hspace{3cm}
Shahjalal University of\\\hspace{3cm}
Science and Technology\\\hspace{3cm}Sylhet, Bangladesh\\\hspace{3cm}
Email: aysha.kamal7@gmail.com}
\and

\IEEEauthorblockN{Md. Saiful Islam}
\IEEEauthorblockA{Department of Computer \\Science and Engineering\\
Shahjalal University of\\
Science and Technology\\Sylhet, Bangladesh\\
Email: saif.acm@gmail.com}}

\IEEEoverridecommandlockouts
\IEEEpubid{\begin{minipage}[t]{\textwidth}\ \\[10pt]
        \centering\normalsize{978-1-7281-5842-6/19/\$31.00~\copyright2019 IEEE}
\end{minipage}}


\maketitle
\begin{abstract}
 Language models are generally employed to estimate the probability distribution of various linguistic units, making them one of the fundamental parts of natural language processing. Applications of language models include a wide spectrum of tasks such as text summarization, translation and classification. For a low resource language like Bengali, the research  in this area so far can be considered to be narrow at the very least, with some traditional count based models being proposed. This paper attempts to address the issue and proposes a continuous-space neural language model, or more specifically an ASGD weight-dropped LSTM language model, along with techniques to efficiently train it for Bengali Language. The performance analysis with some currently existing count based models illustrated in this paper also shows that the proposed architecture outperforms its counterparts by achieving an inference perplexity as low as 51.2 on the held out data set for Bengali. 
\end{abstract}

\hfill \break


\section{Introduction}
Language Models (LMs) can generally be categorized into two variants: continuous-space language models \cite{continuous} and count based language models \cite{countbased}. Traditional statistical models, which constitute of a large proportion of the count-based architectures, follow the general idea of making n-th order Markov assumptions and calculating the n-gram probabilities through the means of counting and subsequent smoothing. Most of the work in Bengali Language has been focused on count based approaches \cite{sabir} and delivered decent performances in tasks such as word clustering. A major drawback of this approach of representing feature spaces through n-gram models is extreme sparsity and often limits the performance of the n-gram models in their applications.  

Despite being a revelation in the art of language modelling, very little work has been done on continuous space language models for Bengali. Variants include the feed-forward neural probabilistic language models (NPLMs) \cite{continuous} and recurrent neural language models (RNNs) \cite{continuous}, which solve the problem of data sparsity that occurs in the traditional n-gram methods. This is done by representing words as vectors (word embeddings) and using them as inputs to a neural language model (NLM). The parameters are learned during the back-propagation phase of the training process. The vectors are created to maintain the property where semantically similar words are kept close to each other in the induced vector space \cite{continuous}.  Neural Language Models have also been used to capture the contextual information at multiple levels including that of sentence, corpus and sub-word.

In this paper, we propose a variant of the recurrent neural language model proposed in \cite{merity}, named Average-Stochastic-Gradient-Descent Weight-Dropped LSTM. We also present a framework that holds multiple techniques to optimize the training of the language model to produce significantly low perplexities on data sets.
This paper follows the structure provided below:
\begin{itemize}
    \item Related Works - This sections provides the necessary background study on some works relevant to this paper. Architectures, strategies and methods described in this section are frequently used in the proposed methodology.
    \item Corpus - The corpus used in the experiments for this paper are summarized in this section.
    \item Methodology - The proposed architecture for language modelling along with the strategies used during the training phase of the neural networks are described in depth.
    \item Experiments - Describes the experimental setup, along with some models used for comparative evaluations.
    \item Results and Discussion - Analysis of the results along with possible reasons are discussed.
    \item Conclusion - The paper concludes with some recommendations and provides scope for future research on this field.
\end{itemize}
\section{Related Works} \label{related}
\subsection{On Bengali Language}
Despite being the $7^{th}$ most spoken language in the world, Bengali suffers from a lack of fundamental resources for natural language processing. Unsurprisingly, the work done so far on language models can be, at the very least, considered unsatisfactory as most of the published researches propose count-based models \cite{sabir}. Although some substantial work was done \cite{bangla_1}\cite{ourpaper} the scarcity of continuous-space models in Bengali language still remains and thus hampers the progress in tasks like text classification, summarization and translation. 

\subsection{On Language Models}
Language Models(LMs) generally come in two major variants, count-based LM and continuous-space LM, along with their own merits and shortcomings.
\newline
\subsubsection{Count-based language models}
Constructing a joint probability distribution of a sequence of words is the fundamental statistical approach to Language Model. n-gram LM model based on Markov assumption can be regarded as an example of this type. In n-gram model, we proceed by predicting one word at a time based on the previous history of the preceding words to get the full sequence. For a preceding sequence of words LM probability is the product of all those word probabilities $p(w_{1},w_{2}, ... ,w_{n})$. Previous m words are considered as history.
$$p( w_{n}  \mid  w_{1} , w_{2} , \ldots , w_{n-1} ) \approx p( w_{n}  \mid  w_{n-m} , \ldots , w_{n-2}, w_{n-1} )$$

This is called Markov chain, and order of the model is the number of previous states(words in this case). The main idea behind n-gram is to predict the probability of $w_{n}$ based on preceding context. If we take only one word $w_{n-1}$ then its called bi-gram. If we divide the frequency of $w_{n-1},w_{n}$ by frequency of $w_{n-1}$ we get the desired result. If we only consider frequency for $w_{n}$ then it would be a uni-gram. The straight forward equation for tri-gram(most often used) can be described as below: $$p(w_{3} \mid w_{1},w_{2}) =  \frac{count(w_{1},w_{2},w_{2})}{ \sum{}_w count(w_{1},w_{2},w)} $$ 

However, if a combination of words is not encountered in the training corpus, a simple n-gram model would predict zero probability for that sequence. But, out-of-sample test cases are likely to occur and assigning zero is not correct. This problem is called sparsity. To counter this problem, various back-off \cite{back-off} and smoothing techniques \cite{smooth1} were introduced, but no good solution exists. \par

Despite the above mentioned smoothing techniques, another drawback of n-gram is the curse of dimensionality. Because, there can be a huge number of possible combination of word sequences and LM has to identify them separately.\par

Besides, n-gram models rely on the exact pattern (i.e. string matching), which is linguistically uninformed. Similarity of different sentences which are syntactically and semantically same can not be provided by n-gram models. Another problem is that we ignore dependency beyond the window. We only take context for n words. So, true conditional probability under Markov assumption is not modeled.
\newline

\subsubsection{Continuous-space language models}
The short-comings of n-gram LM led to the idea of applying Neural Networks and Deep learning on LM so as to be able to extract syntactic and semantic features of languages. \par

This type of language model is also known as neural language model (NLM). There are two main types of NLM: feed-forward neural network based LM for tackling data sparsity and recurrent NN based LM for addressing the problem of limited context. As of late, state-of-the-art performance was achieved by recurrent neural network approaches. Later works are focusing on different (i.e. sub-word level modelling and corpus-level modelling) directions based on RNN and its variant such as LSTM.

\begin{itemize}
    \item Feed-Forward Neural Network Based Models:
\end{itemize} \par
\begin{figure}[h!]
\includegraphics[scale=0.35]{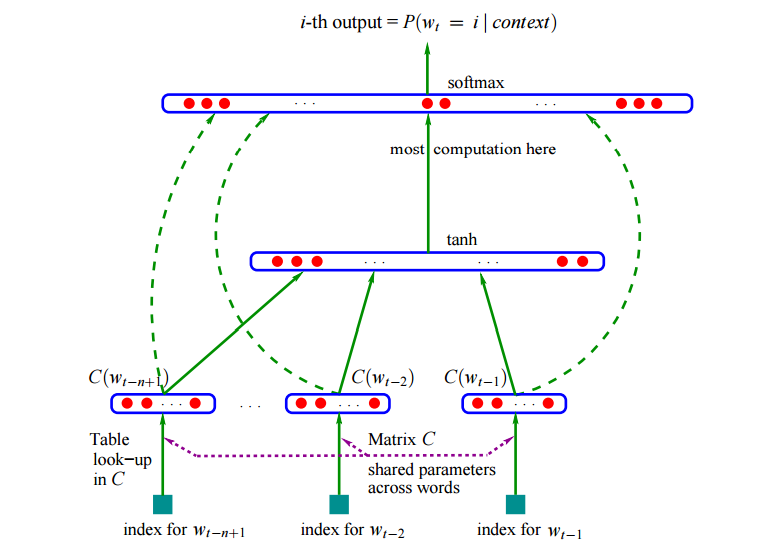}
\centering
\caption{An overview of the network architecture of a neural probabilistic language model \cite{probabilistic_LM} }
\label{Fig:feed_forward}
\end{figure}

Neural probabilistic language model\cite{probabilistic_LM} is the first neural approach to LM. It learns the parameters for conditional probability for next word using a three layer feed-forward NN for previous n-1 words. In figure \ref{Fig:feed_forward} an overview is given for the architecture. 
In the architecture:

\begin{enumerate}
    \item A mapping C is built from each word of vocabulary V to a real-valued, distributed feature vector $C(i) \in R^{m}$, where m is the number of features. C is a matrix of $\mid V\mid \times m $, whose i-th row indicates feature vector of word i.
    
    \item A function g calculates and maps conditional probability of word $w_t$ with context $(C(w_{t-n+1}),\ldots , C(w_{t-1}))$  of the input sequence of feature vectors.
    
    \item Lastly, both the word feature vectors and parameters of probability function is learned with composite function f, where mappings C and g is used.
\end{enumerate}

In this model, we have a distributed word feature vector for each word in the vocabulary. And a function of these feature vector express joint probability function of the words in the sequence. The model then simultaneously learns feature vector and parameters of the probability. Sparseness problem is solved because of the neural network. It also tends to generalize well in comparison to the n-gram model discussed above. However, very long training and testing time is considered a major weakness to this approach. \par

To improve the training and testing speed of NLM, two models were proposed\cite{morin2005hierarchical}\cite{mnih2009scalable}. The basic idea is to cluster words which are similar to reduce calculation and load of the NN. One\cite{morin2005hierarchical} builds a binary hierarchical tree on the vocabulary words using expert knowledge. The other\cite{mnih2009scalable} uses a data-driven method instead of expert knowledge. The best HLBL model\cite{mnih2009scalable} truncates perplexity by 11.1\% compared to the Kneser-Ney smoothed 5-gram LM




\begin{itemize}
    \item Recurrent Neural Network Based Models:
\end{itemize}\par
\begin{figure}[h!]
\includegraphics[scale=0.4]{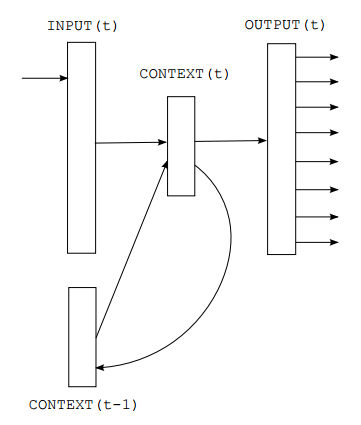}
\centering
\caption{Recurrent neural network model \cite{recurrentLM} }
\label{Fig:rnnlm}
\end{figure}

Unlike feed-forward neural network, we do not use limited size context in recurrent neural networks. When using recurrent connections, information keep circulating inside the networks for arbitrary time as long as needed. We can have better generalization with recurrent neural network based model(RNNLM)\cite{recurrentLM}. Recurrently connected neurons are considered to be short term memory. For a simple architectural overview we can refer to the figure \ref{Fig:rnnlm}. 

    
    
    

For further improvement a variant of RNNLM was presented \cite{RLMextensions} where they implemented factorization on the output layer using classes, which resulted in reducing the computational complexity of the original RNNLM. 
For applications on statistical LM, RNN usually performs better than FNN.

\subsection{On Neural Network Architectures}
\begin{figure}[h!]
\includegraphics[scale=0.5]{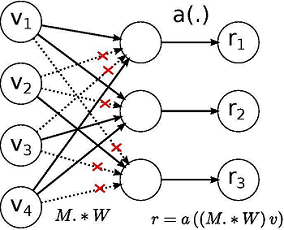}
\centering
\caption{DropConnect Network}
\label{Fig: DropConnect}
\end{figure}

As of recent times, the AWD-LSTM showed excellent performances by reaching the state-of-the-art for language modelling.\cite{AWD}. AWD-LSTM stands for ASGD(Average Stochastic Gradient Descent) Weight-Dropped LSTM. It uses various regularization strategies, which includes DropConnect\cite{dropconnect}, a variant of Average-SGD (NT-ASGD) and some other minor techniques in a very effective way.
\newline
The generic equation for LSTM can be described as follows:
 $$i_{t} =  \sigma ( W^{i} x_{t} +  U^{i} h_{t-1}) $$
 $$f_{t} =  \sigma ( W^{f} x_{t} +  U^{f} h_{t-1})$$
 $$o_{t} =  \sigma ( W^{o} x_{t} +  U^{o} h_{t-1})$$
 $$ \widetilde{c}_{t}  =  tanh ( W^{c}x_{t} + U^{c}h_{t-1}) $$
 $$c_{t} =  i_{t}  \odot  \widetilde{c}_{t} +  f_{t} \odot  +  \widetilde{c}_{t-1}$$
 $$ h_{t} =  o_{t}  \odot  tanh(c_{t})$$
 where, $W^{i}, W^{f}, W^{o}, W^{c}, U^{i}, U^{f}, U^{o}, U^{c}$ represents weight matrices, $x_{t}$ is the vector input to timestep t, $h_{t}$ is the current exposed hidden state, $c_{t}$ is the memory cell state, and $\odot$ is element-wise multiplication.
 \newline
 In conventional LSTM, over-fitting of the RNN creates problem. That is where the DropConnect comes to play. It is applied to randomly selected activation subset on the hidden-to-hidden weight matrices $(U^{i}, U^{f}, U^{o}, U^{c})$ instead of hidden or memory states. This prevents over-fitting without disrupting RNN’s ability to retain long-term dependencies. Figure \ref{Fig: DropConnect} provides an illustration of the DropConnect network.

ASGD is identical with SGD on updating steps, but ASGD also considers previous iterations and return an average value. The variant of ASGD that is used in AWD-LSTM is NT-ASGD(non-monotonically triggered ASGD). In this method ASGD is triggered if validation metric fails to improve for a fixed number of cycles. A constant learning rate is used throughout the experiment.
\newline
Other techniques include variational dropout\cite{variable_dropout} which generates the dropout mask once upon the first call. variable length backpropagation helps reducing the divisibility problem of elements. Embedding Dropout\cite{variable_dropout}, Reduction in Embedding Size, Activation Regularization all resulted in the improvement of the overall structure.
Successful application in achieving state-of-the-art performance on language modeling is what makes it the best choice till date.

\subsection{On Training Neural Networks}

Training neural networks so that the performance culminates as efficiently as possible for the necessary task has become almost an art form in the recent years. This section illustrates some of the techniques that will be employed by this paper.

\begin{figure}[h!]
\includegraphics[scale=0.45]{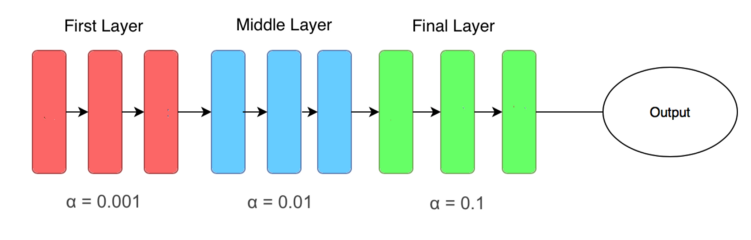}
\centering
\caption{Application of Differential Learning Rates}
\label{diff_learn}
\end{figure}

Learning rates constitute of some of the most important hyper-parameters during the training phase. A technique known as differential learning rates has been seen in numerous researches including the work by \cite{howard}. The implementation of the technique makes sure that higher layers of the neural network train at a faster rate than the deeper ones. Building deep learning models on top of pre-existing elements, such as embedding layers, has delivered excellent performances as of late. The general idea also means that the latter layers will need to be somewhat modified. 
Figure \ref{diff_learn} provides an illustration of differential learning rates \cite{tenthings}.

Another technique introduced by \cite{howard} employs gradual unfreezing of the layers. For instance, in cases of transfer learning, trying to train all the layers of an entire model consisting of pre-trained weights at once creates the risk of catastrophic forgetting of the pre-trained weights. \cite{howard} proposed to gradually unfreeze the layers, starting from the last, which contains the least bit of relevant information initially. 

Despite being one of the most deciding factors in the efficient training of a neural network, finding the right learning rate has been incredibly tedious over the years, often involving multiple trial and error experiments. To address the issue, Leslie Smith published an impressive approach \cite{clr}. The method includes doing a trial run and training the neural network using a low learning rate, but increasing it exponentially with each batch. Meanwhile, for every value of the learning rate, the loss is recorded. The point where the learning rate is highest but the loss is still descending is taken as the optimum learning rate.

\begin{figure}[h!]
\includegraphics[scale=0.4]{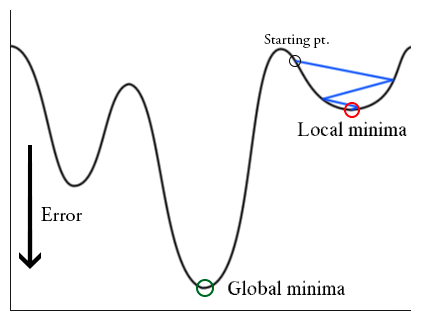}
\centering
\caption{Gradient Descent Stuck at Local Minima \cite{tenthings}}
\label{sgdr_prob}
\end{figure}

\begin{figure}[h!]
\includegraphics[scale=0.35]{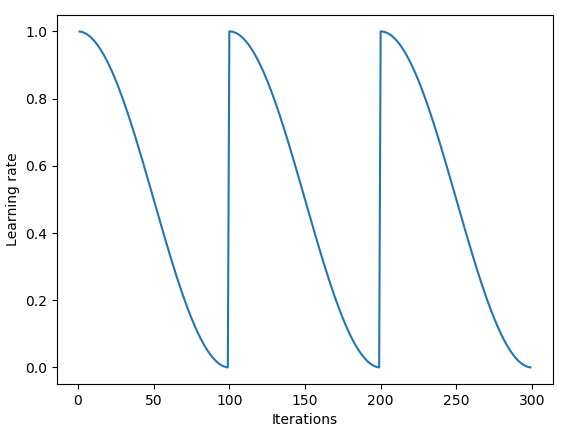}
\centering
\caption{Resetting Learning Rate after each Epoch \cite{tenthings}}
\label{sgdr_solution}
\end{figure}

A common issue during the training phase of deep neural networks is the gradient descent getting stuck at a local minima, instead of reaching the global minimum. \cite{SGDR} suggests that by increasing the learning rate abruptly, the gradient descent can be expected to jump out of the local minima and start searching for the global minimum again. The technique was named stochastic gradient descent and it proved quite effective in the experiments performed by \cite{SGDR}. Figure \ref{sgdr_prob} provides a diagrammatic representation of the issue of local minima. Figure \ref{sgdr_solution} describes how learning rates are restarted after every epoch to avoid the problem.

\section{Corpus}
Using a custom web crawler, we created a large corpus of Bengali Newspaper articles based on 6 topics. Accident, crime, education, entertainment, environment and sports. The training set consists of 10564543 tokens while the test set has 1197254. None of the documents in the test set are present in the training set. The corpus is relatively large compared to the language modelling experiments performed previously for Bengali Language.   

\section{Methodology}
\subsection{Proposed Architecture}

The architecture employed in the neural language model is a variant of the Recurrent Neural Network with LSTM gates\cite{gers}, or more specifically the AWD-LSTM \cite{merity}, described in Section \ref{related}.
The model uses specialized regularization techniques and optimizations that enables high performance for language generation and context generalization. Due to the recurring nature of Bengali language, using the AWD-LSTM RNN seemed like the most appropriate choice. 

\begin{figure}[h!]
\includegraphics[scale=0.6]{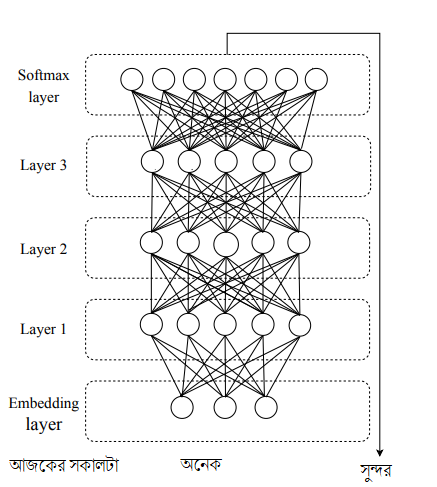}
\centering
\caption{Language Model Architecture}
\label{Fig:awd_bangla}
\end{figure}

The language model mainly consists of 3 intermediate regular LSTM layers with no attention, some short-cut connections and a few other additional mechanisms with numerous drop-out hyper-parameters. The number of hidden activation per layer is 1150. An embedding layer lies beneath, with the chosen embedding size set to 400. A softmax layer is stacked on top of the LSTM layers to provide probabilistic estimates of the next word. The linear decoder is used along with the Adam optimizer and the flattened categorical cross-entropy as the loss function. An illustration of the architecture is provided in figure \ref{Fig:awd_bangla}.

\subsection{Training the language model}

With the maximum vocabulary size set to 60,000,  words that occurred in the training corpus less than 2 times were discarded and replaced with the 'unk' token. Other specialized tokens were added to identify pre-processing methods such as padding(pad), beginning of strings(bos), end of strings(eos) etc. Due to a lack of specialized tokenizers for Bengali Language, space separated words were taken as tokens.
The batch size  was set to 32, while the back-propagation-through-time window was set to 70. The weight decay for all the layers was set to 0.1 and a drop-out multiplier of 0.5 was used for all the LSTM weights to avoid early over-fitting.

\begin{figure}[h!]
\fbox{\includegraphics[scale=0.6]{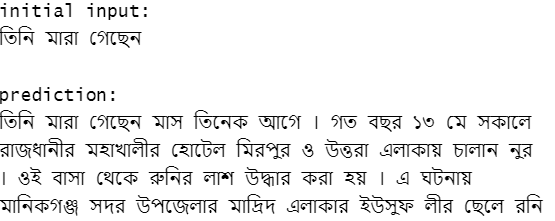}}
\centering
\caption{Example Predictions from the Language Model}
\label{examples}
\end{figure}

With all layers frozen, using the cyclic learning rate (CLR) finder technique \cite{clr}, an appropriate learning rate of 1e-1 was selected and the model was trained for 4 epochs. All the layers were unfrozen next, and as per the differential learning rate technique mentioned in Section \ref{related}, different learning rates were picked for the layers based on their depth for efficient training. A learning rate of 1e-4 was picked for the last layer and 1e-2 was selected for the layer before that. The model was trained for a further 4 epochs, at which point, signs of over-fitting started to occur and the terminal point was called. For every epoch, the technique of stochastic gradient descent with restarts (SGDR) mentioned in Section \ref{related} is applied.
Given a phrase of 3 initial words, the language model at this point was capable of completing entire paragraphs in Bengali language as illustrated in figure \ref{examples}.

\section{Experiments}
To analyze the performance of the proposed framework, we attempt to recreate some previously proposed models and see how they fair in terms of perplexity on the held out data set. Perplexity is a measurement of how well a probability model predicts a test data. In the context of Natural Language Processing, perplexity is one way to capture the degree of 'uncertainty' a model has in predicting (assigning probabilities to) some text. Low perplexity is good and high perplexity is bad since the perplexity is the exponentiation of the entropy. The models used for performance analysis are as follows: 
\subsection{Bi-gram language model}
 
 We draw our first comparison with the very famous and traditionally used bi-gram language model, much similar to the work done in \cite{JelMer}. Variants of this model has been seen to be used repeatedly for modelling Bengali Language. Suppose $q_{t} =  l(freq( w_{t-1},w_{t-2} ))$ represents the frequency of occurrence of the window $(w_{t-1},w_{t-2})$. The conditional probability, thereby, take the form of a conditional mixture:
 $$ P ( w_t  | w_{t-1}  ) =   \alpha _{0}( q_t ) p_0 +
 \alpha _1( q_t ) p_1 ( w_t ) +  \alpha _2( q_t ) p_2 ( w_t | w_{t-1}  )
 $$
 
 with conditional weights $\alpha _{i}( q_{t} ) \geq  0 ,  \sum{}_i \alpha _{i}( q_{t} ) = 1  $.
 The base predictors here are as follows:
 $p_{0} = 1 /|V| , p_{1}$ is a uni-gram and $p_{2}(i|j)$ is the bi-gram . We keep rest of the experimental measures much similar to the models proposed by \cite{sabir}. 
 


\subsection{LSTM and CNN language models}
We also draw comparisons with a variation of a continuous space neural language models to see how the proposed framework, with it's own modifications, compares against simpler models. 
We train an LSTM based language model (Simple LSTM) following much of the conventions by \cite{zaremba}. The model has two layers of LSTMs that are unrolled for 10 epochs. Hidden states are initialized to 0. The batch size if kept 32. The number of activation per layer is chosen to be 200. An initial learning rate of 1 is used. \par 
We also try a character level CNN (Character CNN) language model that accepts a sequence of encoded characters  as input, much similar to the architecture used for text classification by \cite{character}. The alphabet used in our model consists of all bengali letters, 10 bengali digits, 25 other characters including the new line. The model is 3 layers deep with 2 convolutional and 1 fully connected layer. The input features equal to 85 due to our character count and the input feature length is set to 1000. The rest of the parameters are kept consistent with the model used in \cite{character}.

\section{Results and Discussion}

The perplexities of the models achieved on the test set of the corpus are summarized in table \ref{resulttable}.

\begin{table}[h!]
\centering
\caption{Perplexity Comparison of the Language Models}
\begin{tabular}{||c c||} 
 \hline
 Model & Perplexity on Test Set\\ [0.5ex] 
 \hline\hline
 AWD-LSTM (Proposed) & 51.2\\ 
 Simple LSTM & 227\\
 Character CNN & 125\\ 
 Bi-gram & 860.1\\[1ex] 
 \hline
 
\end{tabular}
\break

\label{resulttable}
\end{table}

The diagrammatic representation provides us with a clear view that the proposed architecture outperforms all the other models, some by an impressive margin.  The Simple LSTM architecture, which was used as a control, achieves a perplexity of 227 while the character level CNN language models seems to perform slightly better, achieving 125. The bi-gram model did not seem to work well for the corpus, with a perplexity of 860.1, which highlights the weaknesses of n-gram models mentioned in Section \ref{related}. The AWD-LSTM LM achieves a perplexity as low as 51.2. This improvement over the other models may be due the employment of careful regularization and dropouts through out the intermediate layers, which prevent over-fitting and shift the model towards better generalization and less memorization. The training strategies used in the AWD LSTM which are absent in the other models also seem to compliment the architecture for a performance boost.

\section{Conclusion}
In this work, we discuss strategies for effective training of continuous-space neural language models and introduce the weight-dropped LSTM to Bengali Language. On the corpora consisting of Bengali news articles, the proposed approach yields a perplexity of 51.2 which is significantly lower than the perplexities achieved through traditional language models currently existing in Bengali. We believe that, the core reason for these improvements over the pre-existing models is that the proposed architecture employs DropConnect mask on the hidden-to-hidden weight matrices, as a means to prevent over-fitting across the recurrent connections, thus being able to generalize much better than most continuous-space neural models. Since the applications of language models have flourished over the years, we hope to see the implementation of this model for tasks such as text summarization, classification and translation of Bengali language in the future.

\section{Acknowledgement}
This paper owes its gratitude to the Department of Computer Science and Engineering, Shahjalal University of Science and Technology (SUST) and SUST NLP Research Group for providing us with the right facilities and research environment.  All authors of this paper have major contributions in the research and rightfully deserve equal credit.

\vspace{12pt}

\bibliography{paper} 
\bibliographystyle{ieeetr}
\end{document}